\documentclass[preprint,12pt]{elsarticle}

\makeatletter
\def\ps@pprintTitle{%
\let\@oddhead\@empty
\let\@evenhead\@empty
\let\@oddfoot\@empty
\let\@evenfoot\@empty
}
\makeatother

\usepackage{amssymb}

\usepackage{amsmath}

\usepackage{subcaption}
\usepackage{booktabs}
\usepackage{multirow}
\usepackage{float}
\usepackage{comment}

\usepackage{color}
\definecolor{red}{rgb}{1,0,0}
\definecolor{blue}{rgb}{0,0,1}
\usepackage{ulem}

\begin{document}

\begin{frontmatter}



\title{T-Graph: Enhancing Sparse-view Camera Pose Estimation by Pairwise Translation Graph}


\author[inst1]{Qingyu Xian}
\author[inst2]{Weiqin Jiao}
\author[inst2]{Hao Cheng}
\author[inst1]{Berend Jan van der Zwaag} 
\author[inst1]{Yanqiu Huang\corref{cor1}}
\affiliation[inst1]{organization={Pervasive Systems Research Group, Faculty of Electrical Engineering, Mathematics and Computer Science, University of Twente}, 
            city={Enschede},
            country={The Netherlands}
            }
\affiliation[inst2]{organization={Department of Earth Observation Science, Faculty of Geo-Information Science and Earth Observation (ITC), University of Twente},city={Enschede},country={The Netherlands}}
\cortext[cor1]{Corresponding author. Email: yanqiu.huang@utwente.nl}

\begin{abstract}
Sparse-view camera pose estimation, which aims to estimate the 6-Degree-of-Freedom (6-DoF) poses from a limited number of images captured from different viewpoints, is a fundamental yet challenging problem in remote sensing applications.  
Existing methods often overlook the translation information between each pair of viewpoints, leading to suboptimal performance in sparse-view scenarios.
To address this limitation, we introduce T-Graph, a lightweight, plug-and-play module to enhance camera pose estimation in sparse-view settings. 
T-graph takes paired image features as input and maps them through a Multilayer Perceptron (MLP). 
It then constructs a fully connected translation graph, where nodes represent cameras and edges encode their translation relationships.
It can be seamlessly integrated into existing models as an additional branch in parallel with the original prediction, maintaining efficiency and ease of use. 
Furthermore, we introduce two pairwise translation representations, \textit{relative-t} and \textit{pair-t}, formulated under different local coordinate systems. While \textit{relative-t} captures intuitive spatial relationships, \textit{pair-t} offers a rotation-disentangled alternative. 
The two representations contribute to enhanced adaptability across diverse application scenarios, further improving our module's robustness.  
Extensive experiments on two state-of-the-art methods (RelPose++ and Forge) using public datasets (C03D and IMC PhotoTourism) validate both the effectiveness and generalizability of T-Graph. 
The results demonstrate consistent improvements across various metrics, notably camera center accuracy, which improves by 1\% to 6\% from 2 to 8 viewpoints. 

\end{abstract}




\end{frontmatter}


\section{Introduction}
\label{sec:intro} 
Multi-view camera pose estimation is a fundamental task in computer vision. It involves estimating the 6-Degree-of-Freedom (6-DoF) poses (i.e., translation and rotation in 3D space) of cameras given an unordered set of images captured from different viewpoints.
Sparse-view camera pose estimation is a more challenging subset of the general multi-view pose estimation task, where the goal is to infer or optimize the camera pose corresponding to each view under the condition of having only a limited number of viewpoints available.
In the field of remote sensing, camera pose estimation plays a crucial role in orthorectification, multi-view image fusion, and multi-temporal image registration \cite{wang2018deep,yu2021universal}. Furthermore, it is widely applied in Simultaneous Localization and Mapping (SLAM) \cite{CHEN2022209, WANG202435, YAO2024198} and 3D reconstruction \cite{stucker2022resdepth, yu2021automatic, li2022ransac, gao2023general} based on remote sensing imagery. In disaster monitoring and structural health monitoring (SHM) \cite{worden2007fundamental}, unmanned aerial vehicles (UAVs) are commonly employed for image acquisition, where camera pose estimation enables 3D change detection based on imagery.


Traditional camera pose estimation methods are primarily based on multi-view geometry, such as Structure-from-Motion (SfM) \cite{sturm2012benchmark, schonberger2016structure}. 
These methods are theoretically grounded, independent from labeled data, can provide explainable results, and offer superior accuracy under ideal conditions.
However, they are highly sensitive to texture variations, feature mismatches, and challenging conditions such as wide-baseline or sparse-view scenarios. 
Particularly, in sparse-view scenarios, overlapping regions between viewpoints become limited, reducing the effectiveness of geometric constraints and making traditional SfM methods unreliable. 

In contrast, deep learning-based methods exhibit greater robustness by learning data-driven priors from similar distributions. 
They leverage deep neural networks to directly predict camera poses or feature correspondences, enabling better generalization across diverse environmental conditions and improved handling of textureless or repetitive regions. 
The integration of architectures such as Convolutional Neural Networks (CNNs) \cite{kehl2017ssd, xiang2017posecnn}, Transformers \cite{lin2024RelPose++, sinha2023sparsepose}, and diffusion models \cite{wang2023posediffusion, zhang2024cameras} further enhances their robustness and accuracy. 
However, existing deep learning approaches for \( n \) sparse input images typically regress or generate \( n \) corresponding rotation matrices and translation vectors, 
while neglecting the valuable correlation information inherent in each paired viewpoint, resulting in limited performance under sparse-view scenarios. 
Specifically, these methods commonly assume a fixed world origin (usually placed at the first frame) and model the translation of each camera relative to this origin, which limits the exploitation of global inter-camera relationships, a critical shortcoming in sparse-view scenarios.

To address these limitations, we incorporate each paired translation supervision to better exploit the inter-camera relationship.
Our method is driven by two key motivations: 
First, in sparse-view scenarios, the limited number of viewpoints leads to severe information sparsity, making it challenging for the network to extract enough reliable correlations. 
By explicitly modeling pairwise translations, which encode the relative translation between each pair of viewpoints, the model can effectively leverage all available inter-camera relationships to enrich the scene understanding. 
Second, unlike existing methods that rely solely on camera-to-origin translations, our method constructs a fully connected graph, where nodes represent cameras and edges capture pairwise translations. 
This formulation introduces global information that enables the network to better perceive the overall spatial configuration of the camera system, thereby enhancing pose estimation accuracy in sparse-view settings. 
To realize this pairwise translation supervision, we design a lightweight, plug-and-play MLP-based module, called T-Graph, that predicts the translation graph from pairwise image features.
This T-Graph can be seamlessly integrated into existing end-to-end camera pose estimation frameworks. 
By sharing feature extractors and jointly optimizing parameters, our module introduces an additional constraint that complements existing methods and improves pose estimation performance. 

\begin{figure}[tb]
  \centering
  \includegraphics[scale=0.6]{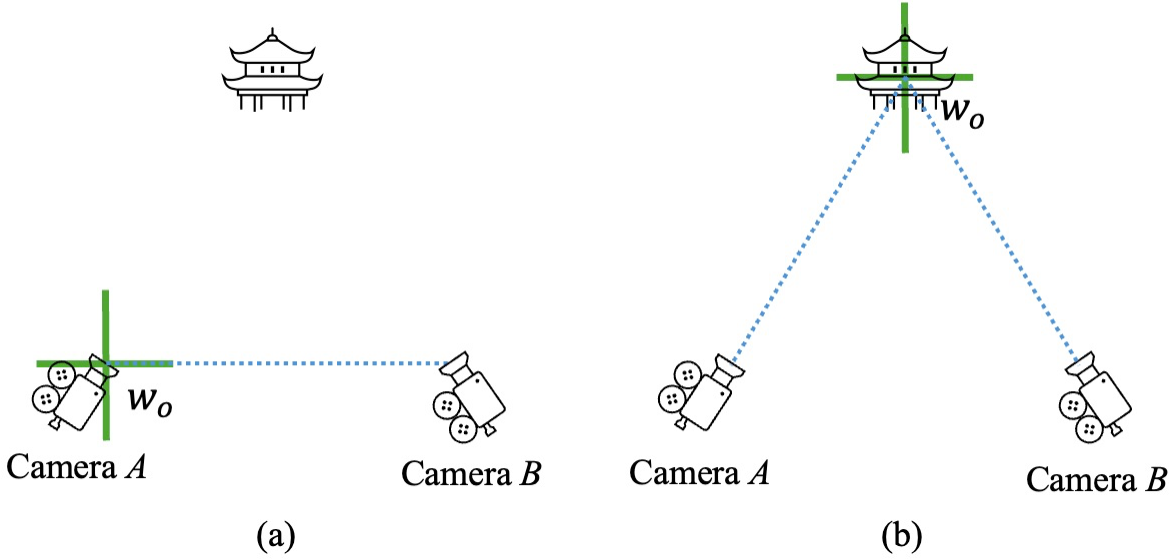}
  \caption{Illustration of two pairwise translation representations based on different coordinate systems. The green cross indicates the origin of the world coordinate system. (a) Coordinate system of \textit{relative-t}, (b) Coordinate system of \textit{pair-t}.}
  \label{fig: two coordinates}
\end{figure}

Furthermore, we propose two types of pairwise translation to cope with various camera scenarios.
Namely, the relative translation (termed \textit{relative-t}) between two cameras at different locations expresses the position of camera \textit{B} relative to camera \textit{A}, where camera \textit{A} is treated as the world origin termed $w_o$, as shown in Fig.~\ref{fig: two coordinates}(a).
\textit{relative-t} works well even if camera orientations are approximately parallel, which is often the case in sparse-view camera pose estimation.
Alternatively, we define the intersection point of two cameras’ optical axes as the world origin ($w_o$) and express the translation of each camera as the location of $w_o$ in each camera’s coordinate frame, as shown in Fig.~\ref{fig: two coordinates}(b).
In this way, we decouple translation from rotation via a common intersection point, making the learning task of camera pose estimation more efficient when the optical axes of each camera pair approximately intersect at a common point.

To validate our approach, we conducted extensive comparative experiments based on two state-of-the-art methods (RelPose++ \cite{lin2024RelPose++} and Forge \cite{jiang2024few}) and two publicly available datasets (C03D(v2) \cite{reizenstein2021common} and IMC PhotoTourism \cite{jin2021image}). 
RelPose++ adopts an energy-based generative scheme specifically for rotation estimation, while Forge is a purely discriminative model. These two methods cover two distinct and representative paradigms in learning-based camera pose estimation. 
The two chosen public datasets exhibit significant differences: the first primarily consists of ``object-centered" everyday objects, while the second encompasses tourist landmarks captured from a wide range of viewpoints. 
The results demonstrate that our enhancement module consistently boosts pose estimation performance across various methods and datasets, while also offering valuable insights into the role of pairwise translation constraints in improving camera pose accuracy. Our main contributions are summarized as follows:
\begin{itemize}
\item We introduce a novel, lightweight plug-and-play module, termed the T-graph that formulates pairwise translation as a fully connected graph to enhance camera pose estimation in sparse-view scenarios.  
\item We propose two pairwise translation representations, \textit{relative-t} and \textit{pair-t}, each tailored to different camera configurations. \textit{relative-t} is particularly well-suited for scenarios where the majority of camera orientations are nearly parallel, whereas \textit{pair-t}, which is rotation-disentangled, is more appropriate for configurations where camera rays are approximately co-planar and exhibit clear convergence
\item Extensive experiments on state-of-the-art methods (RelPose++ and Forge) across diverse datasets (CO3D and IMC PhotoTourism) demonstrate that our module delivers notable performance improvements, confirming the effectiveness of pairwise translation constraints in enhancing camera pose estimation.

\end{itemize}

\section{Related work}
This paper primarily focuses on multi-view camera pose estimation. Accordingly, the existing methods can be broadly categorized into two main groups: geometry-based pose estimation and learning-based pose estimation. In this section, we review and discuss representative approaches from both categories.

\vspace{3pt}
\noindent\textbf{Geometry-Based Pose Estimation.}
The classical SfM \cite{schonberger2016structure} algorithm is a technique for recovering camera poses and 3D scenes from an unordered set of images, primarily relying on feature matching, geometric constraints, and optimization algorithms. The general workflow includes: extracting keypoints and feature descriptors using algorithms such as Scale-Invariant Feature Transform (SIFT) \cite{lowe2004distinctive} and performing feature matching across different images; computing the essential or fundamental matrix and recovering the relative pose by filtering out outliers via RANSAC \cite{fischler1981paradigm}; reconstructing 3D points through triangulation \cite{hartley2003multiple}; and optimizing camera poses, either incrementally or globally, followed by Bundle Adjustment (BA) \cite{triggs2000bundle} to minimize the re-projection error. 
Recent work \cite{huang2024bundle} has introduced novel motion-based geometric constraints to enable accurate reconstruction and pose estimation in uncalibrated, unsynchronized, and non-overlapping camera setups, thereby transforming low-cost consumer-grade multi-camera data into high-quality 3D models. 
Furthermore, methods like SuperPoint \cite{detone2018superpoint} and SuperGlue \cite{sarlin2020superglue} have significantly improved the accuracy of feature extraction and matching, and have been integrated into the SfM pipeline to substantially enhance pose estimation and reconstruction results. 
However, in sparse-view scenarios, the limited performance of feature matching and the lack of sufficient viewpoint constraints may lead to drift or even convergence failure in SfM solutions. 

\vspace{3pt}
\noindent\textbf{Learning-Based Pose Estimation.} 
Compared to geometry-based methods, learning-based approaches are more suitable for camera pose estimation across diverse environments, as they do not rely on discrete feature points or feature matching.
For general multi-view camera pose estimation, early CNN-based methods \cite{kehl2017ssd,xiang2017posecnn} directly regress 6-DoF poses from RGB images using convolutional architectures, but often suffer from limited accuracy and generalization due to the expressive power limitations of neural networks.
Hybrid architectures like TransCNNLoc \cite{tang2024transcnnloc} integrate CNNs, Swin Transformers \cite{liu2021swin}, and dynamic object recognition to enhance feature robustness and attain centimeter-level accuracy of pose estimation, yet they do not explicitly model pairwise correspondences across viewpoints, limiting their potential in sparse-view scenarios.
DiffPoseNet \cite{parameshwara2022diffposenet} integrates optical flow estimation within a deep neural network, introducing a normal flow-based camera pose estimation method. It designs the NFlowNet network to learn the normal flow and employs a differentiable cheirality constraint layer for end-to-end optimization. 
The underlying motivation of this method aligns with that of our proposed module, as both are designed to refine camera poses estimated by existing approaches, though they are implemented from different perspectives.
Diffusion models, which have demonstrated remarkable performance in generative modeling, have also been applied to camera pose estimation. Posediffusion \cite{wang2023posediffusion} utilizes a Denoising Diffusion Probabilistic Model (DDPM) \cite{ho2020denoising} to perform forward noise addition and iteratively refine predictions toward the correct solution, integrating epipolar constraints during the prediction phase. However, the performance of this method remains limited in sparse scenes.

In contrast to general multi-view settings with abundant views from varying perspectives, sparse-view scenarios introduce additional challenges: the overlapping regions between adjacent viewpoints become more limited, and the available input information is significantly reduced. These factors collectively pose substantial difficulties for accurate camera pose estimation.
To overcome the additional challenges, \cite{jin2021planar} explores the planar information available in such settings and proposes a method that simultaneously estimates camera poses and reconstructs the planar surfaces of indoor scenes. Sparsepose \cite{sinha2023sparsepose} first regresses an initial camera pose, followed by iterative refinement using a sampling-based autoregressive approach. FORGE \cite{jiang2024few} is designed with two branches that separately extract 2D and 3D features, which are then fused to solve for the camera pose and subsequent 3D reconstruction.
RelPose \cite{zhang2022RelPose} employs an energy-based model to characterize the distribution of relative rotations from a set of cameras, thereby enabling joint inference over multiple images to obtain consistent camera rotations. 
By modeling the relative rotations among all viewpoints, the extraction and utilization of effective features are enhanced. But this method is limited to predicting rotation only.
Building on RelPose, RelPose++ \cite{lin2024RelPose++} introduces a Transformer architecture to incorporate feature information from viewpoints other than the current one, and further proposes a novel global coordinate system to reduce the impact of ambiguity in rotation on translation estimation, resulting in more robust pose predictions. 
\cite{cerkezi2024sparse} proposes a hybrid method for 3D object reconstruction from sparse 360° views that combines a mesh-guided sampling scheme with a neural surface representation, achieving state-of-the-art results.
However, these methods still fall short in thoroughly exploiting the information between pairwise viewpoints, which can result in suboptimal performance, particularly in challenging scenarios with limited inputs or minimal overlap between the reference frame and other captured images.
Therefore, in this work, we aim to enhance the performance of camera pose estimation in sparse-view scenarios. To this end, we propose T-Graph, which is designed to fully exploit pairwise translation information and improve the model's ability to perceive informative features.

\section{Methodology}
In this section, we first present how T-Graph operates within an end-to-end camera pose estimation pipeline in Sec \ref{sec:reform}. 
We then provide an explanation of two pairwise translation representations in Sec. \ref{sec:arc}.
The detailed learning objectives are specified in Sec. \ref{sec:loss}.

\begin{figure}[t!]
  \centering
  \includegraphics[width=1\linewidth]{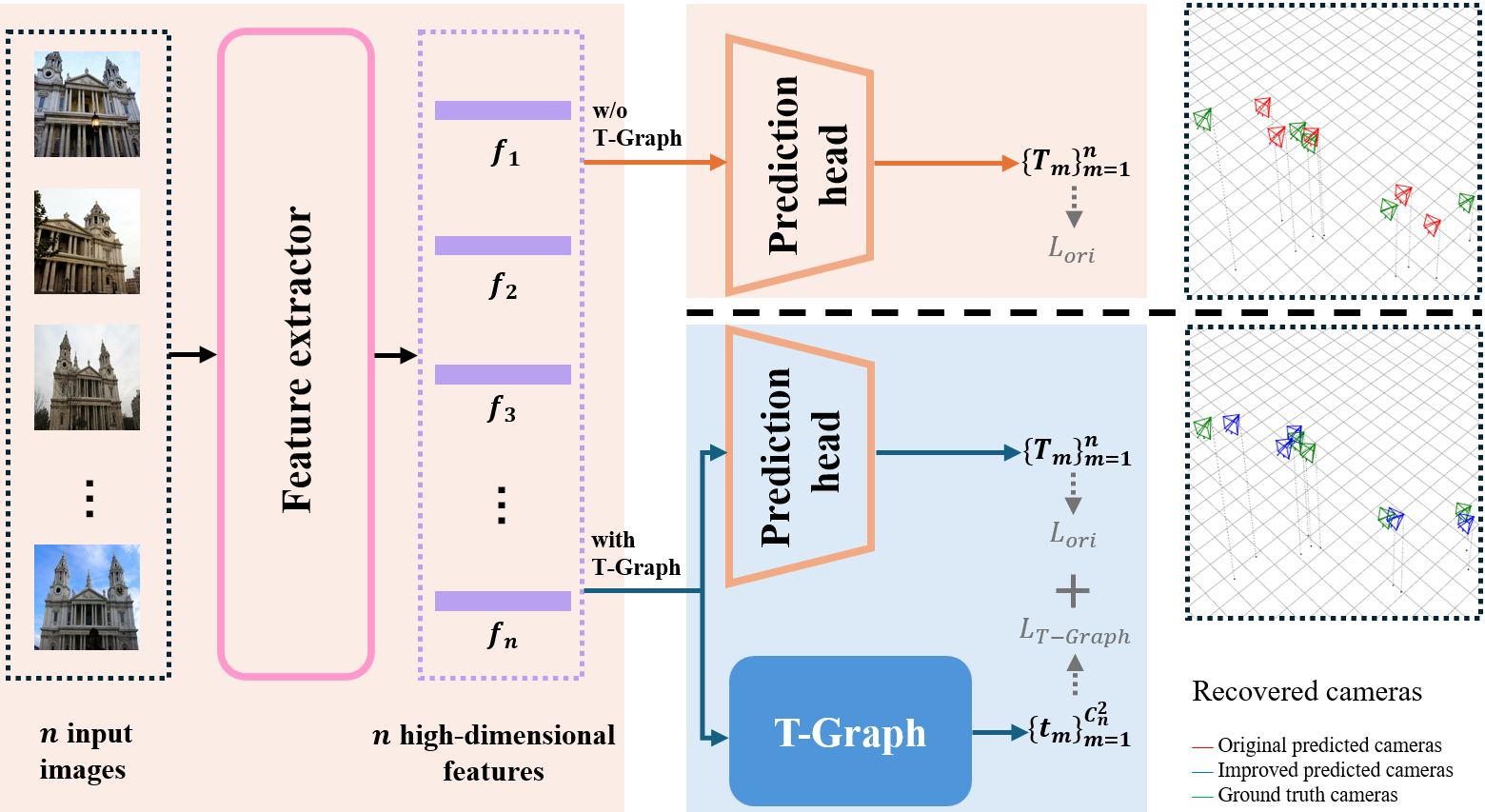}
  \caption{Comparison of camera pose estimation architectures with and without T-Graph.  
  $\left\{ T_m \right\}_{m=1}^{n}$ represents the output (rotation, translation) of the baseline model and $\left\{ t_m \right\}_{m=1}^{C_n^2}$ represents the output (pairwise translation) of T-Graph. 
  The pink region illustrates the baseline model structure without T-Graph, while the blue region shows the modified model structure after introducing T-Graph, which works as a parallel prediction branch to assist learning. Note that the T-Graph branch is only active during training for loss optimization and is removed during inference.}
  \label{fig: architecture}
\end{figure}

\subsection{Design of T-Graph}
\label{sec:reform}

T-Graph can be seamlessly integrated into commonly used camera pose estimation networks by introducing a new branch parallel to the baseline model’s prediction head, as shown in Fig.~\ref{fig: architecture}. Sharing the same input features and upstream parameters, T-Graph provides additional supervision during training, guiding the feature extractor to learn more discriminative and globally informative representations related to camera poses, and thereby enhancing the overall accuracy of pose estimation.

\begin{figure}[tb]
  \centering
  \includegraphics[width=1\linewidth]{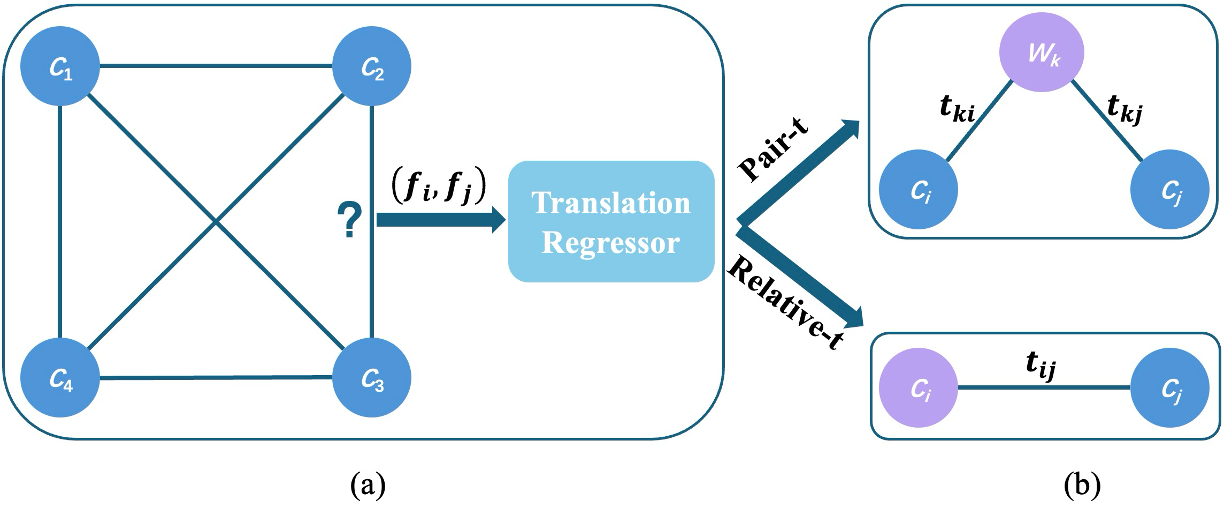}
  \caption{T-Graph (simplified with four cameras) module with two different pairwise translation representations. (a) T-Graph, (b) Two types of pairwise translation representations.
  }
  \label{fig: translation graph}
\end{figure}

As detailed in Fig.~\ref{fig: translation graph}(a), the proposed T-Graph is a complete graph where each node represents a camera $C_i$, and each edge models the translation relationship between paired cameras through a dedicated translation regressor. Specifically, we employ a lightweight MLP as the translation regressor. Under $n$ sparse viewpoints (ranging from 2 to 8), the T-Graph module takes a pair of high-dimensional features $(f_i, f_j)$ as input, which are extracted by the feature extractor of the baseline model from the images captured by the $i$-th and $j$-th cameras, respectively. 
The module then outputs the translation relationship between these paired cameras, i.e., \text{T-Graph}($f_i$, $f_j$), which is modeled by a shared translation regressor. 
In total, there are $C_n^2$ such pairwise translation relationships, all processed by the same regressor. 

Given that T-Graph models only the translation relationship between camera pairs, rather than absolute translations with respect to a fixed world origin, its outputs are not adopted as final predictions. 
Instead, they function as an additional supervision signal to effectively guide the baseline model in learning more discriminative features and improving pose estimation accuracy. 
Notably, T-Graph operates exclusively during training as a plug-and-play enhancement module and is omitted from the final model at inference, thereby preserving the original inference efficiency of the baseline model. 
Regarding the choice of baseline models, T-Graph is conceptually compatible with a wide range of learning-based camera pose estimation methods. In this study, we validate its effectiveness on two representative methods in Sec. \ref{sec:experiment}.

\subsection{Two Pairwise Translation Representations}
\label{sec:arc}
To model pairwise translations within T-Graph, we propose two different representations, \textit{relative-t} and \textit{pair-t}, each formulated under a distinct local coordinate system. 
These representations are designed to accommodate diverse application scenarios and enhance the flexibility and robustness of the model.  


Figure~\ref{fig: translation graph}(b) denotes the incorporation of the two pairwise translation representations into our T-Graph. 
More specifically, for \textit{relative-t}, the translation along each edge of T-Graph is defined as $t_{ij}$, representing the relative translation between two cameras, $C_i$ and $C_j$, as shown in the lower part of Fig.~\ref{fig: translation graph}(b). 
For \textit{pair-t}, the translation on each edge is then defined as $t_{ki}$ and $t_{kj}$, where $k$ represents the intersection point of the optical axes of the two cameras, serving as the world origin $W_{k}$, as illustrated in the upper part of Fig.~\ref{fig: translation graph}(b). 
It is important to note that the intersection point refers to the location in 3D space that minimizes the distance to both optical axes.
Accordingly, $t_{ki}$ and $t_{kj}$ denote the locations of $W_{k}$ in the coordinate frames of $C_i$ and $C_j$, respectively.

We next provide an explanation of the principles underlying the two types of pairwise translation representations.
To represent pairwise translation, a natural way is to define the relative translation (\textit{relative-t}) between two cameras at different locations by expressing the position of $C_j$ relative to $C_i$, where $C_i$ is treated as the world origin, as shown in the lower part of Fig.~\ref{fig: translation graph}(b).
Formally, the \textit{relative-t} can be written as:
\begin{equation}
\label{eq:relative t}
    t_{i\rightarrow j} = X_j^c - R_{i\rightarrow j} \cdot X^w,
\end{equation}
where $R_{i\rightarrow j}$ is the rotation matrix of $C_j$ relative to $C_i$,
$X^w$ is the point in the world frame, and $X_j^c$ is the corresponding point in the camera frame of $C_j$. 
While intuitive, this representation remains entangled with camera rotation. Since $t_{i\rightarrow j}$ depends on $R_{i\rightarrow j}$, the model must implicitly infer rotation in order to learn translation, introducing an additional learning burden. 

To disentangle translation from rotation, a novel pairwise translation formulation, \textit{pair-t} is proposed, as illustrated in the upper part of Fig.~\ref{fig: translation graph}(b), where $W_{k}$ is placed at the intersection point of the optical axes of the two cameras.
In this case, $X^w$ = (0, 0, 0), $C_i$'s translation is equal to $X_i^c$ = (0, 0, $D_i^c$) and $C_j$'s is equal to $X_j^c$ = (0, 0, $D_j^c$), where $D_i^c$ and $D_j^c$ denote the distances from $C_i$ and $C_j$ to the world origin, respectively. 
This formulation decouples translation and rotation, eliminating the influence of rotation and thus simplifying the learning task.

These two representation schemes are not inherently superior or inferior to one another; rather, they are suited to different application scenarios. 
The \textit{pair-t} representation assumes that the optical axes of each camera pair converge at an approximate point, making it particularly suitable for scenarios where cameras are nearly co-planar and clearly convergent. For example, in the CO3D dataset \cite{reizenstein2021common}, cameras are generally oriented toward the target center, aligning well with this assumption. 
In such cases, \textit{pair-t} enables an effective decoupling of rotation and translation, thereby reducing the learning complexity and enhancing model performance. 
In contrast, datasets like IMC PhotoTourism \cite{jin2021image} present camera distributions that diverge from the above assumption. 
Here, most camera optical axes are approximately parallel or converge at very small angles, making the estimation of intersection points in \textit{pair-t} unstable, thereby degrading performance. 
Therefore, under such conditions, \textit{relative-t} becomes a more appropriate choice for representing pairwise translations. 

\subsection{Learning objective}
\label{sec:loss}
During data preprocessing, we normalize the ground-truth pairwise translations, either $(t_{ki}, t_{kj})$ or $t_{ij}$, for each sequence to facilitate stable model convergence. Specifically, given a set of input images, we compute the $\mathcal{L}_2$ norm of each translation vector, identify the maximum norm within the sequence, and normalize all $(t_{ki}, t_{kj})$ or $t_{ij}$ vectors by dividing them by this maximum value.

As shown in Eq.~\eqref{eq:loss_t_graph}, we adopt an $\mathcal{L}_1$ loss between the predicted T-Graph and the corresponding ground truth. 
\begin{equation}
\label{eq:loss_t_graph}
    \mathcal{L}_{\text {T-Graph }} = \left\{
    \begin{array}{ll}
        k_1 \sum \left\| \text{T-Graph}\left(f_i, f_j\right) - \left(t_{k i}, t_{k j}\right) \right\|_1, & \text{for \textit{pair-t}} \\
        k_2 \sum \left\| \text{T-Graph}\left(f_i, f_j\right) - t_{i j} \right\|_1, & \text{for \textit{relative-t}}
    \end{array}
    \right.
\end{equation}
To balance this loss with the translation loss from the main prediction branch of the baseline model, we introduce scaling factors $k_1$ and $k_2$ for \textit{pair-t} and \textit{relative-t}, respectively. 
In the main prediction branch, the translation is estimated for each of the $n$ viewpoints relative to a fixed world origin, resulting in $n$ loss terms. 
In contrast, T-Graph models pairwise translations: specifically, \textit{relative-t} involves $C_n^2$ terms, while \textit{pair-t} involves $2 \times C_n^2$ terms due to two translation vectors per camera pair. 
To prevent this loss component from dominating the overall model training, we scale it by a coefficient such that the number of outputs from the T-Graph module is consistent with the number of viewpoints. 
Accordingly, we define $k_1$ and $k_2$ as follows to ensure that the magnitudes of the two losses remain comparable:
\begin{equation}
k_1 = \frac{n}{2 \times C_n^2}, \quad
k_2 = \frac{n}{C_n^2}
\end{equation}

The overall loss function of the model is formulated in Eq.~\eqref{eq:loss_all}, which consists of the original loss $\mathcal{L}_{\text{ori}}$ including rotation loss and translation loss of the baseline model, along with the additional T-Graph loss $\mathcal{L}_{\text{T-Graph}}$. 
\begin{equation}
\label{eq:loss_all}
\mathcal{L}_{\text{full}} = \mathcal{L}_{\text{ori}} + \mathcal{L}_{\text{T-Graph}}
\end{equation}

During training, the model is iteratively optimized under the joint supervision of T-Graph loss and the original loss of the baseline model.

\section{Experiment}
\label{sec:experiment}
In this section, we provide a detailed description of the experimental setup, followed by both quantitative and qualitative results. 
\subsection{Experimental setup}
In this subsection, we present the experimental setup, including the datasets, the baseline models, the ablation studies, and the evaluation metrics.

\noindent\textbf{Dataset.} 
We evaluate the proposed method on two datasets: CO3D \cite{reizenstein2021common} and IMC PhotoTourism \cite{jin2021image}. 
The two datasets differ significantly in object categories and camera distributions, making them suitable for evaluating the generalizability of our proposed module across varying scenarios. 

CO3D consists of video sequences spanning 51 object categories, with ground-truth camera poses annotated using COLMAP \cite{schonberger2016structure}. In each sequence, the camera follows a motion trajectory that approximately revolves around the target object. Following the experimental setup of RelPose++ \cite{lin2024RelPose++}, we trained on data from 41 categories (training set) and validated the effectiveness of the proposed method on the remaining 10 categories (test set).

IMC PhotoTourism contains image data of over 20 renowned landmarks worldwide, collected from user-captured photos on Flickr. The ground truth camera poses for this dataset were also derived from SfM reconstructions using COLMAP. According to the official publicly released dataset split, we trained on data from 10 scenes (training set) and validated the proposed method on the other 8 scenes (test set).

\begin{figure}[tb]
  \centering
  \includegraphics[width=1\linewidth]{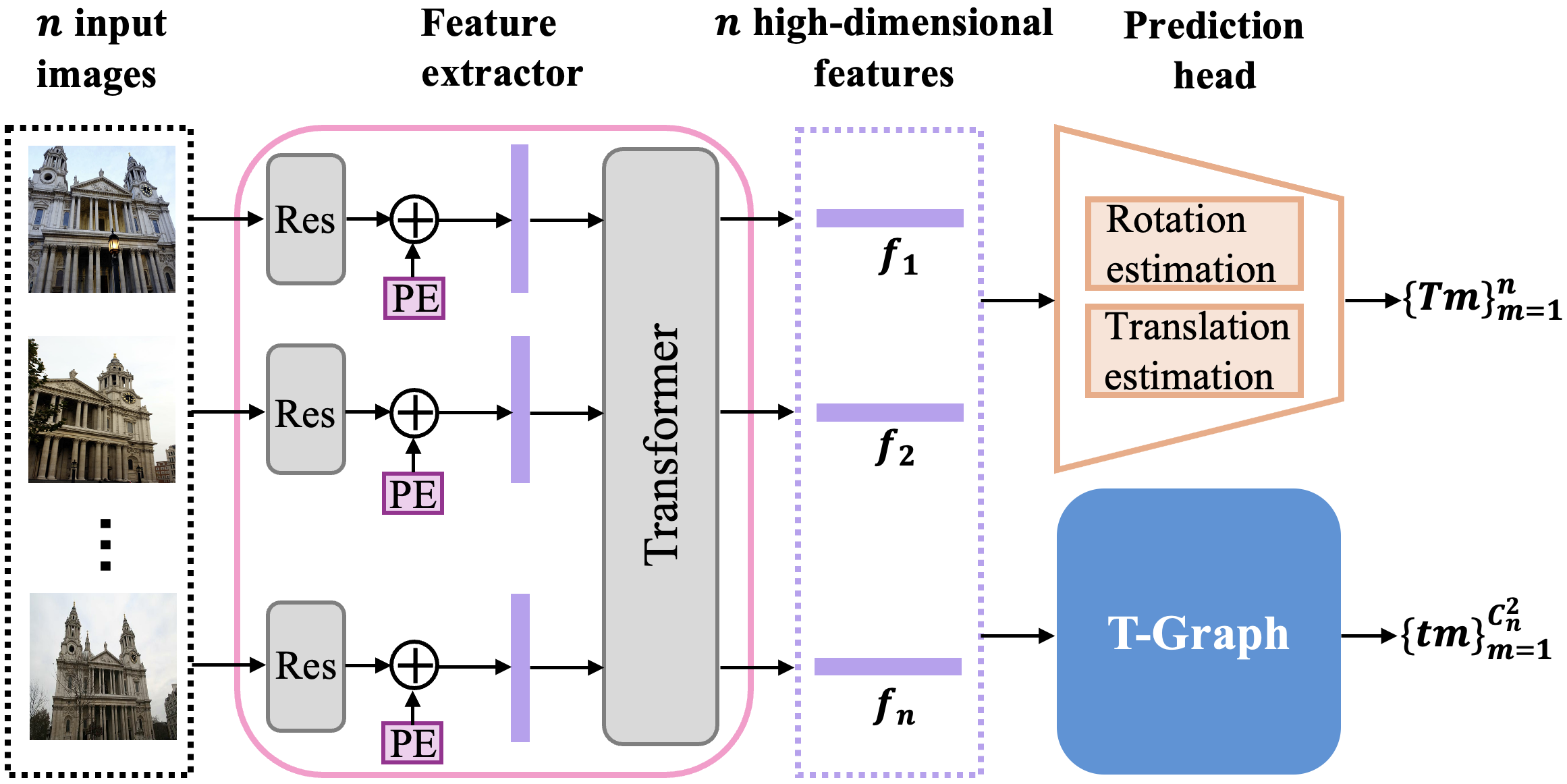}
  \caption{RelPose++ \cite{lin2024RelPose++} combined with T-Graph. RelPose++ adopts a ResNet-50 backbone to extract image features, which are then fused with positional embeddings and bounding box parameters and fed into a Transformer network. This method directly regresses camera translations while modeling the distribution of rotations through an energy-based approach.}
  \label{fig: RelPose++ with t graph}
\end{figure}

\begin{figure}[tb]
  \centering
  \includegraphics[width=1\linewidth]{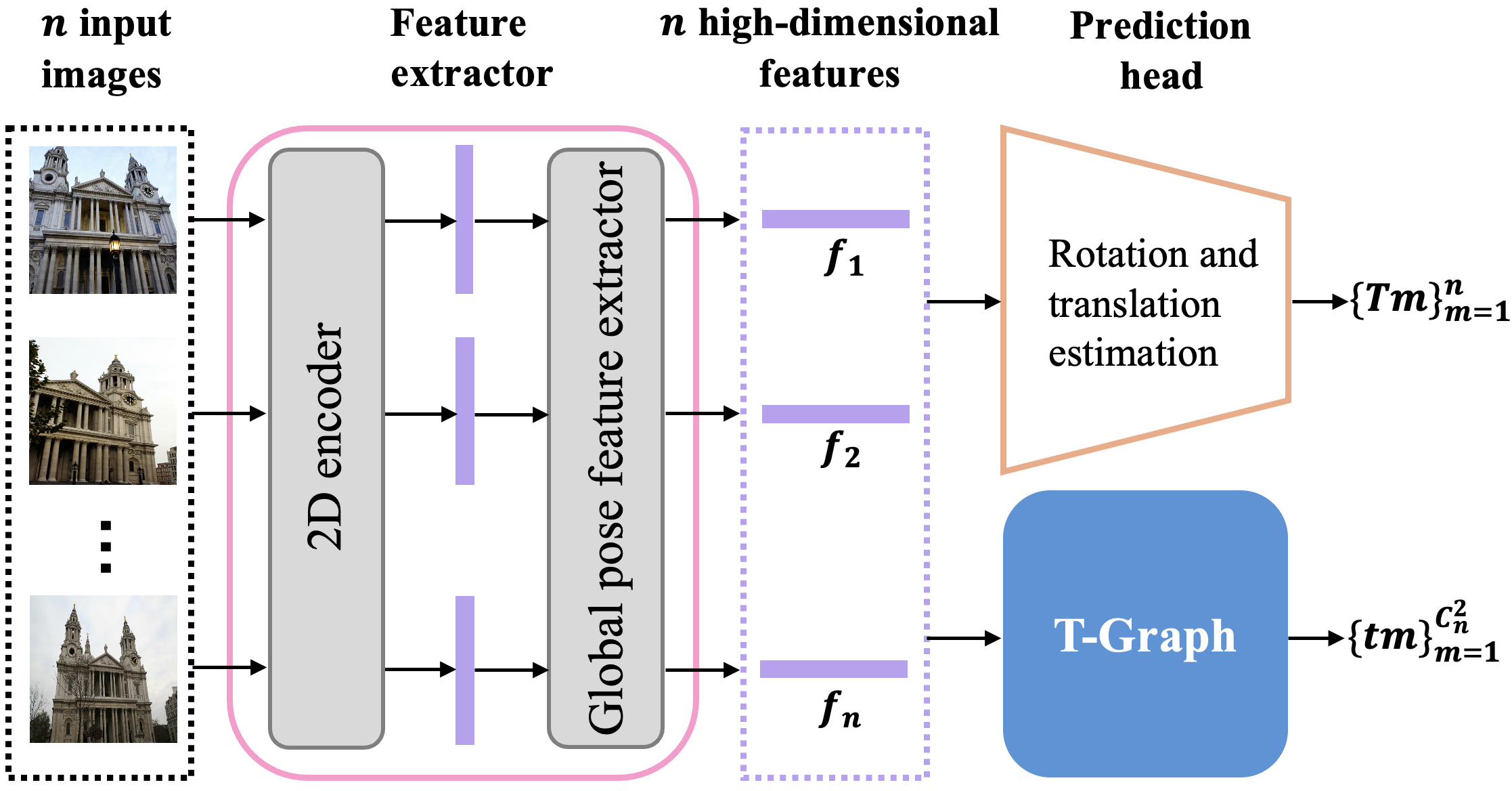}
  \caption{Forge-2D \cite{jiang2024few} combined with T-Graph. Forge-2D employs a multi-layer CNN as a 2D encoder to extract image features. These features are then passed to a global pose feature extractor, which integrates cross-attention and self-attention mechanisms to capture pose-related information. Finally, an MLP jointly regresses camera rotations and translations in a unified manner.}
  \label{fig: forge with t graph}
\end{figure}

\noindent\textbf{Baseline models and ablation studies.} 
Regarding the selection of the baseline model, we first adopt RelPose++ \cite{lin2024RelPose++}, which incorporates a generative, energy-based model for final rotation regression. 
To complement this, we further include Forge \cite{jiang2024few} as a second baseline, as it is a purely discriminative approach for camera pose estimation. 
Since Forge consists of both 2D and 3D branches, and our study focuses on RGB image-only inputs, we restrict our experiments to its 2D branch. 
To distinguish it from the original Forge model, we refer to it as Forge-2D. 
This setup allows us to evaluate T-Graph across two representative frameworks, thereby demonstrating its potential to generalize to a wide range of end-to-end camera pose estimation methods. 
In both methods, the feature extractor processes each input image captured by an individual camera to obtain its corresponding image feature. 
The integration of T-Graph into these baseline models follows a unified strategy: It is appended after the feature extractor, running parallel to the original prediction head, as shown in Figs.~\ref{fig: RelPose++ with t graph} and \ref{fig: forge with t graph}. 
Notably, Forge models translation by treating the camera of the first frame as the world origin. 
After extracting $k$ image features, Forge applies cross-attention between the first image feature and each of the remaining $k-1$ features to generate $k-1$ pose features, which are subsequently fed into the prediction head for pose estimation, as illustrated in Fig.~\ref{fig: forge with t graph}. 
To utilize all $k$ image features in a unified manner, we apply self-attention to the first image feature to produce its corresponding pose feature, so that all $k$ pose features can serve as inputs to T-Graph. 
For each method, the inference and post-processing stages remain unchanged from the baseline models after incorporating T-Graph.
In the T-Graph module, to balance computational cost and model performance, a uniform 6-layer MLP architecture was adopted across all experiments.

To fully evaluate the effectiveness and generalizability of T-Graph with two pairwise translation representations, we conducted four groups of comparative experiments across different models and datasets, comprising a total of 11 complete training sessions, as summarized in Table \ref{tab:exp_display}. Notably, the evaluation results of RelPose++ on CO3D are directly taken from its original publication \cite{lin2024RelPose++}, and thus, no additional training was conducted in our experiments.

In each comparative experiment across Experiment Groups (EG) 1 to 4, we maintained consistent experimental settings to ensure the fairness and validity of the comparisons. Specifically, for each model in EG 1, training was performed on a single H100 GPU with a batch size of 22, while keeping all other hyperparameters identical to those used in RelPose++. 
In EG 2 to EG 4, all models were trained on a single A100 GPU due to the unavailability of H100 GPUs. The batch sizes were set to 8, 16, and 16, and the learning rates to 1e-4, 1e-5, and 1e-4, respectively.
Across all experiments, the batch size was carefully selected to maximize GPU memory utilization and computational efficiency.
For models trained on the CO3D dataset, we follow the learning rate setting of 1e-5 as used in RelPose++. For models trained on the smaller-scale IMC PhotoTourism dataset, we found that a higher learning rate of 1e-4 led to faster and more stable convergence.
Additionally, the AdamW optimizer was employed throughout, and early stopping was applied to prevent overfitting and improve training stability.

\begin{table}[]
\caption{Comparative experiments}
\centering
\fontsize{8pt}{8pt}\selectfont
\label{tab:exp_display}
\begin{tabular}{@{}llccccccc@{}}
\toprule
Method                     & Dataset                             & T-Graph & Experiment Group   No.   \\ \toprule
\multirow{7}{*}{RelPose++} & \multirow{3}{*}{CO3D}           & pair   T            & \multirow{3}{*}{1} \\ \cmidrule(l){3-3}
                           &                                     & relative   T        &                    \\ \cmidrule(l){2-4} 
                           & \multirow{4}{*}{IMC   PhotoTourism} & W/O                 & \multirow{4}{*}{2} \\ \cmidrule(l){3-3}
                           &                                     & pair   T            &                    \\ \cmidrule(l){3-3}
                           &                                     & relative   T        &                    \\ \midrule
\multirow{10}{*}{Forge-2D} & \multirow{4}{*}{CO3D}           & W/O                 & \multirow{4}{*}{3} \\ \cmidrule(l){3-3}
                           &                                     & pair   T            &                    \\ \cmidrule(l){3-3}
                           &                                     & relative   T        &                    \\ \cmidrule(l){2-4} 
                           & \multirow{4}{*}{IMC   PhotoTourism} & W/O                 & \multirow{4}{*}{4} \\ \cmidrule(l){3-3}
                           &                                     & pair   T            &                    \\ \cmidrule(l){3-3}
                           &                                     & relative   T        &                    \\ \bottomrule
\end{tabular}
\end{table}

\noindent\textbf{Evaluation metrics.}
Similar to RelPose++, the input consists of 2 to 8 images randomly sampled from a sequence, and each method outputs the corresponding 6-DoF camera pose ({$R_m$, $t_m$}). To accurately assess the performance variation of each method before and after incorporating T-Graph, we report three metrics proposed in RelPose++: rotation accuracy, camera center accuracy, and translation accuracy. 
And we use the same threshold for each metric as in RelPose++.
All metrics remain invariant under the global similarity transformation between the predicted and ground truth cameras.

\noindent\textit{Rotation Accuracy.}
We evaluate the relative rotation errors between each prediction and its corresponding ground truth, and report the proportion of cases where the error is within 15 degrees.

\noindent\textit{Camera Center Accuracy.}
Also referred to as camera localization error, this metric is widely used in standard benchmarks within the SLAM \cite{sturm2012benchmark} community. However, because the predicted camera center and the ground truth camera center may reside in different coordinate systems, a direct comparison is not feasible. 
Therefore, following RelPose++, we first compute the optimal similarity transform between the two sets using a Least-Squares \cite{umeyama1991least} approach to align them. 
We then report the proportion of aligned predicted camera centers that fall within 20\% of the scene scale relative to the ground truth camera centers, where the scene scale is defined as the distance from the centroid of the ground truth camera centers to the farthest camera. 
Thus, the evaluation threshold corresponds to 20\% of this scene scale.

\noindent\textit{Translation Accuracy.}
The evaluation of translation accuracy follows a procedure similar to that of camera center accuracy. We first compute the optimal similarity transform between the predicted translations and the ground truth translations to align them. Subsequently, we report the proportion of aligned predicted translations that are within 20\% of the scene scale relative to the ground truth translations, with the scene scale defined as previously described.

To ensure a fair comparison with the published results of RelPose++ on the CO3D dataset, the first set of experiments adopted the sequence order file provided by RelPose++, thereby guaranteeing identical input configurations. For viewpoint counts ranging from 2 to 8, five independent sampling trials were conducted following the same protocol as RelPose++, and the average performance across these trials was reported. In the subsequent experiments, to alleviate computational overhead during test, following \cite{zhang2022RelPose}, a fixed random seed was employed, and a single random sampling trial was performed for each viewpoint count within the same range.

\subsection{Quantitative results}
Table \ref{tab:exp1} presents the results of camera center accuracy, rotation accuracy, and translation accuracy for models trained on CO3D, including RelPose++, RelPose++ with \textit{pair-t}, and RelPose++ with \textit{relative-t}, under 2 to 8 viewpoints. 
Notably, the camera center accuracy at a threshold of 0.2 is always 1 when the number of views is 2, due to the use of a global similarity transformation between the predicted and ground truth cameras. Therefore, it is unnecessary to report or compare the results under this setting. 
The results show that incorporating \textit{pair-t} exhibits notable improvements in all metrics over the baseline model, while incorporating \textit{relative-t} shows notable improvements in most metrics. 
This indicates that T-Graph facilitates the optimization of rotation, translation, and camera center estimation, resulting in a global parameter optimization and overall performance improvement. 
Furthermore, we observe that the improvements brought by \textit{relative-t} are less stable and less pronounced than those from \textit{pair-t}, suggesting that the \textit{pair-t}-based representation is more suitable for the CO3D dataset. 
The experimental results here are consistent with our theoretical analysis. In the CO3D dataset, the camera viewpoints are distributed around the target objects in a roughly circular manner, and the optical axes of any two cameras tend to converge at a common point. 
Under such conditions, using \textit{pair-t} enables T-Graph to decouple rotation and translation, facilitating more efficient and accurate learning of translation within T-Graph. 
The optimization of T-Graph in turn promotes the optimization of the shared network parameters, ultimately enhancing the performance of the baseline model in both rotation and translation prediction. 

\begin{table}[htbp]
\centering
\caption{Results of EG 1 based on RelPose++: Camera Center Accuracy at 0.2, Rotation Accuracy at 15° and Translation Accuracy at 0.2 on CO3D. Best values are highlighted in boldface, and second-best values are underlined.}
\label{tab:exp1}
\fontsize{8pt}{8pt}\selectfont
\begin{tabular}{@{}l|l|ccccccc@{}}
\toprule
Metric                                                                              & \# of Images     & 2                    & 3              & 4              & 5              & 6              & 7              & 8              \\ \midrule
\multirow{3}{*}{\begin{tabular}[c]{@{}l@{}}Camera Center\\Acc at 0.2\end{tabular}} & RelPose++        & -                    & 0.825          & 0.756          & 0.719          & 0.699          & 0.685          & 0.675          \\ \cmidrule(l){2-9} 
                                                                                     & Ours(pair-t)     & -                    & \textbf{0.848} & \textbf{0.778} & \textbf{0.748} & \textbf{0.730} & \textbf{0.712} & \textbf{0.702} \\
                                                                                     & Ours(relative-t) & -                    & \underline{0.834}          & \underline{0.762}          & \underline{0.732}          & \underline{0.720}          & \underline{0.705}          & \underline{0.701}          \\ \midrule
\multirow{3}{*}{\begin{tabular}[c]{@{}l@{}}Rotation\\  Acc at 15°\end{tabular}}      & RelPose++        & \underline{0.698}    & \underline{0.711}  & 0.719             & 0.728             & 0.738             & 0.744             & 0.749          \\ \cmidrule(l){2-9} 
                                                                                     & Ours(pair-t)     & \textbf{0.699}       & \textbf{0.719}     & \textbf{0.735}    & \textbf{0.753}    & \textbf{0.757}    & \textbf{0.765}    & \textbf{0.769} \\
                                                                                     & Ours(relative-t) & 0.684                & 0.701              & \underline{0.722} & \underline{0.735} & \underline{0.748} & \underline{0.754} & \underline{0.758}          \\ \midrule
\multirow{3}{*}{\begin{tabular}[c]{@{}l@{}}Translation\\  Acc at 0.2\end{tabular}}    & RelPose++       & \underline{0.960}    & \underline{0.938}  & 0.931             & 0.923             & 0.922             & 0.918             & 0.916          \\ \cmidrule(l){2-9} 
                                                                                     & Ours(pair-t)     & \textbf{0.966}       & \textbf{0.946}     & \textbf{0.936}    & \textbf{0.934}    & \textbf{0.928}    & \textbf{0.924}    & \textbf{0.923} \\
                                                                                     & Ours(relative-t) & 0.958                & 0.937              & \underline{0.933} & \underline{0.930} & \underline{0.927} & \underline{0.923} & \underline{0.921}          \\ \bottomrule
\end{tabular}
\end{table}

\begin{table}[hhhh]
\centering
\caption{Results of EG 2 based on RelPose++: Camera Center Accuracy at 0.2, Rotation Accuracy at 15° and Translation Accuracy at 0.2 on IMC PhotoTourism. Best values are highlighted in boldface and second-best values are underlined.}
\label{tab:exp2}
\fontsize{8pt}{8pt}\selectfont
\begin{tabular}{@{}l|l|ccccccc@{}}
\toprule
Metric                                                                              & \# of Images     & \multicolumn{1}{c}{2} & \multicolumn{1}{c}{3} & \multicolumn{1}{c}{4} & \multicolumn{1}{c}{5} & \multicolumn{1}{c}{6} & \multicolumn{1}{c}{7} & \multicolumn{1}{c}{8} \\ \midrule
\multirow{3}{*}{\begin{tabular}[c]{@{}l@{}}Camera Center\\Acc at 0.2\end{tabular}} & RelPose++         & -                     & 0.585                 & 0.406                 & 0.346                 & 0.414                 & \underline{0.377}     & 0.369                 \\ \cmidrule(l){2-9} 
                                                                                     & Ours(pair-t)     & -                     & \underline{0.596}    & \underline{0.411}     & \underline{0.354}     & \textbf{0.425}        & \textbf{0.395}        & \textbf{0.386}        \\
                                                                                     & Ours(relative-t) & -                     & \textbf{0.605}        & \textbf{0.416}        & \textbf{0.359}        & \underline{0.423}    & \textbf{0.395}        & \underline{0.380}                 \\ \midrule
\multirow{3}{*}{\begin{tabular}[c]{@{}l@{}}Rotation\\Acc at 15°\end{tabular}}      & RelPose++        & \underline{0.627}       & 0.616                 & 0.614                 & 0.624                 & 0.623                 & 0.614                 & 0.613                 \\ \cmidrule(l){2-9} 
                                                                                     & Ours(pair-t)     & 0.624                 & \underline{0.642}     & \underline{0.625}     & \underline{0.636}     & \underline{0.631}     & \underline{0.628}     & \underline{0.625}                 \\
                                                                                     & Ours(relative-t) & \textbf{0.641}        & \textbf{0.643}        & \textbf{0.638}        & \textbf{0.641}        & \textbf{0.639}        & \textbf{0.639}        & \textbf{0.633}        \\ \midrule
\multirow{3}{*}{\begin{tabular}[c]{@{}l@{}}Translation\\Acc at 0.2\end{tabular}}    & RelPose++        & \textbf{0.595}        & 0.376                 & \underline{0.295}      & \underline{0.278}     & 0.357                 & 0.332                 & \underline{0.354}                 \\ \cmidrule(l){2-9} 
                                                                                     & Ours(pair-t)     & \underline{0.589}    & \underline{0.386}     & 0.293                 & \underline{0.278}      & \underline{0.367}     & \underline{0.346}     & 0.349                 \\
                                                                                     & Ours(relative-t) & \underline{0.589}    & \textbf{0.390}        & \textbf{0.303}        & \textbf{0.292}        & \textbf{0.377}        & \textbf{0.360}        & \textbf{0.363}        \\ \bottomrule
\end{tabular}
\end{table}

To verify whether the aforementioned pattern applies to other datasets, we further conducted EG 2 on IMC PhotoTourism. 
Table \ref{tab:exp2} shows the results of camera center accuracy, rotation accuracy, and translation accuracy for models trained on IMC PhotoTourism, including Relpose++, RelPose++ with \textit{pair-t}, and RelPose++ with \textit{relative-t}, under 2 to 8 viewpoints. 
It can be seen that the T-Graph module consistently improves the performance of the baseline model. 
However, in contrast to EG 1, the improvements brought by \textit{pair-t} are less stable and less substantial than those achieved by \textit{relative-t}, indicating that the \textit{relative-t} representation is more suitable for IMC PhotoTourism. 
This finding also aligns with our initial design, as the camera configurations in IMC PhotoTourism differ a lot from those in CO3D. 
Specifically, cameras rarely face the target center in IMC PhotoTourism. 
Instead, many camera optical axes are approximately parallel or converge at small angles. 
Consequently, employing \textit{pair-t} here causes some challenge to model optimization, due to the instability and large variance in the estimated intersection points. 
In contrast, \textit{relative-t} provides a more reliable representation of the translation relationships between camera pairs in this scenario. 

\begin{table}[hhhh]
\centering
\caption{Results of EG 3 based on Forge-2D: Camera Center Accuracy at 0.2, Rotation Accuracy at 15° and Translation Accuracy at 0.2 on CO3D. Best values are highlighted in boldface and second-best values are underlined.}
\label{tab:exp3}
\fontsize{8pt}{8pt}\selectfont
\begin{tabular}{@{}l|l|ccccccc@{}}
\toprule
Metric                                                                              & \# of Images     & \multicolumn{1}{c}{2} & \multicolumn{1}{c}{3} & \multicolumn{1}{c}{4} & \multicolumn{1}{c}{5} & \multicolumn{1}{c}{6} & \multicolumn{1}{c}{7} & \multicolumn{1}{c}{8} \\ \midrule
\multirow{3}{*}{\begin{tabular}[c]{@{}l@{}}Camera Center\\Acc at 0.2\end{tabular}} & Forge-2D        & -                     & 0.613                 & \underline{0.480}                 & 0.386                 & 0.363                 & 0.334                 & 0.328                 \\ \cmidrule(l){2-9} 
                                                                                     & Ours(pair-t)     & -                     & \textbf{0.659}        & \textbf{0.512}        & \textbf{0.434}        & \textbf{0.408}        & \textbf{0.397}        & \textbf{0.362}        \\
                                                                                     & Ours(relative-t) & -                     & \underline{0.618}                 & \underline{0.480}                 & \underline{0.428}                 & \underline{0.385}                 & \underline{0.376}                 & \underline{0.354}                 \\ \midrule
\multirow{3}{*}{\begin{tabular}[c]{@{}l@{}}Rotation\\Acc at 15°\end{tabular}}      & Forge-2D        & \underline{0.707}                 & 0.618                 & 0.588                 & 0.545                 & 0.521                 & 0.518                 & \underline{0.508}                 \\ \cmidrule(l){2-9} 
                                                                                     & Ours(pair-t)     & \textbf{0.711}        & \underline{0.630}                 & \textbf{0.608}        & \underline{0.546}                 & \textbf{0.541}        & \textbf{0.544}        & \textbf{0.529}        \\
                                                                                     & Ours(relative-t) & 0.703                 & \textbf{0.632}        & \underline{0.598}                 & \textbf{0.548}        & \underline{0.534}                 & \underline{0.532}                 & 0.507                 \\ \midrule
\multirow{3}{*}{\begin{tabular}[c]{@{}l@{}}Translation\\Acc at 0.2\end{tabular}}   & Forge-2D        & 0.250                 & 0.363                 & \underline{0.358}                 & 0.357                 & 0.357                 & 0.344                 & 0.360                 \\ \cmidrule(l){2-9} 
                                                                                     & Ours(pair-t)     & \underline{0.257}                 & \textbf{0.408}        & \textbf{0.367}        & \textbf{0.390}        & \textbf{0.405}        & \textbf{0.384}        & \textbf{0.393}        \\
                                                                                     & Ours(relative-t) & \textbf{0.268}        & \underline{0.386}                 & 0.352                 & \underline{0.366}                 & \underline{0.381}                 & \underline{0.381}                 & \underline{0.363}                 \\ \bottomrule
\end{tabular}
\end{table}

\begin{table}[hhhh]
\centering
\caption{Results of EG 4 based on Forge-2D: Camera Center Accuracy at 0.2, Rotation Accuracy at 15° and Translation Accuracy at 0.2 on IMC PhotoTourism. Best values are highlighted in boldface, and second-best values are underlined.}
\label{tab:exp4}
\fontsize{8pt}{8pt}\selectfont
\begin{tabular}{@{}l|l|ccccccc@{}}
\toprule
Metric                                                                              & \# of Images     & \multicolumn{1}{c}{2} & \multicolumn{1}{c}{3} & \multicolumn{1}{c}{4} & \multicolumn{1}{c}{5} & \multicolumn{1}{c}{6} & \multicolumn{1}{c}{7} & \multicolumn{1}{c}{8} \\ \midrule
\multirow{3}{*}{\begin{tabular}[c]{@{}l@{}}Camera Center\\Acc at 0.2\end{tabular}} & Forge-2D        & -                     & 0.336                 & 0.234                 & 0.231                 & 0.236                 & 0.259                 & 0.276                 \\ \cmidrule(l){2-9} 
                                                                                     & Ours(pair-t)     & -                     & \underline{0.342}        & \underline{0.243}        & \underline{0.242}        & \underline{0.246}        & \underline{0.261}        & \underline{0.282}        \\
                                                                                     & Ours(relative-t) & -                     & \textbf{0.343}        & \textbf{0.254}        & \textbf{0.246}        & \textbf{0.259}        & \textbf{0.280}        & \textbf{0.286}        \\ \midrule
\multirow{3}{*}{\begin{tabular}[c]{@{}l@{}}Rotation\\Acc at 15°\end{tabular}}      & Forge-2D        & 0.788                 & 0.700                 & 0.661                 & 0.631                 & 0.609                 & 0.570                 & 0.574                 \\ \cmidrule(l){2-9} 
                                                                                     & Ours(pair-t)     & \underline{0.801}        & \textbf{0.730}        & \underline{0.688}        & \underline{0.662}                 & \underline{0.635}        & \underline{0.613}        & \underline{0.611}        \\
                                                                                     & Ours(relative-t) & \textbf{0.802}        & \underline{0.727}        & \textbf{0.694}        & \textbf{0.673}        & \textbf{0.638}        & \textbf{0.620}        & \textbf{0.619}        \\ \midrule
\multirow{3}{*}{\begin{tabular}[c]{@{}l@{}}Translation\\Acc at 0.2\end{tabular}}   & Forge-2D        & 0.136                 & 0.130                 & 0.144                 & 0.169                 & 0.186                 & 0.210                 & 0.231                 \\ \cmidrule(l){2-9} 
                                                                                     & Ours(pair-t)     & \underline{0.146}                 & \underline{0.140}        & \underline{0.148}        & \textbf{0.187}        & \underline{0.201}        & \underline{0.228}        & \underline{0.239}        \\
                                                                                     & Ours(relative-t) & \textbf{0.163}        & \textbf{0.143}        & \textbf{0.156}        & \underline{0.186}                 & \textbf{0.208}        & \textbf{0.236}        & \textbf{0.247}        \\ \bottomrule
\end{tabular}
\end{table}

To further evaluate the generalization of T-Graph, we repeated the procedures of EG 1 and 2 using Forge-2D, resulting in EG 3 and 4, which correspond to Table~\ref{tab:exp3} and Table~\ref{tab:exp4}, respectively. 
The experimental results clearly indicate that T-Graph consistently enhances the performance of the baseline model across two different datasets. On CO3D, the improvement brought by \textit{pair-t} surpasses that of \textit{relative-t}, whereas the opposite trend is observed on IMC PhotoTourism. 
Overall, the results exhibit trends consistent with those observed in the corresponding experiments with RelPose++. 

Through four sets of comparative experiments, we demonstrate that T-Graph consistently brings performance gains across two distinct methods and two different datasets, suggesting its potential generalizability to a wide range of approaches and application scenarios. Moreover, these experiments validate the suitability of the two proposed pairwise translation representations under different camera configurations. 

In addition, Table \ref{tab:param_comparison} presents the model parameters of the two methods, RelPose++ and Forge-2D, before and after the integration of T-Graph, along with the corresponding changes. It can be observed that the additional model parameters introduced by T-Graph are minimal, indicating that T-Graph is a lightweight augmentation module.
\begin{table}[ht]
    \centering
    \caption{Comparison of model weight sizes (in megabytes) before and after using T-Graph.}
    \label{tab:param_comparison}
    \fontsize{8pt}{8pt}\selectfont
    \begin{tabular}{lccc}
        \toprule
        \textbf{Method} & \textbf{Params (MB)} & \textbf{Params w/ T-Graph (MB)} & \textbf{$\Delta$ Params (\%)} \\
        \midrule
        RelPose++  & 512 & 537 (+25) & +5 \\
        Forge-2D   & 148 & 164 (+16) & +11 \\
        \bottomrule
    \end{tabular}
\end{table}

\subsection{Qualitative results}
\begin{figure}[H]
  \centering
  \includegraphics[width=1\linewidth]{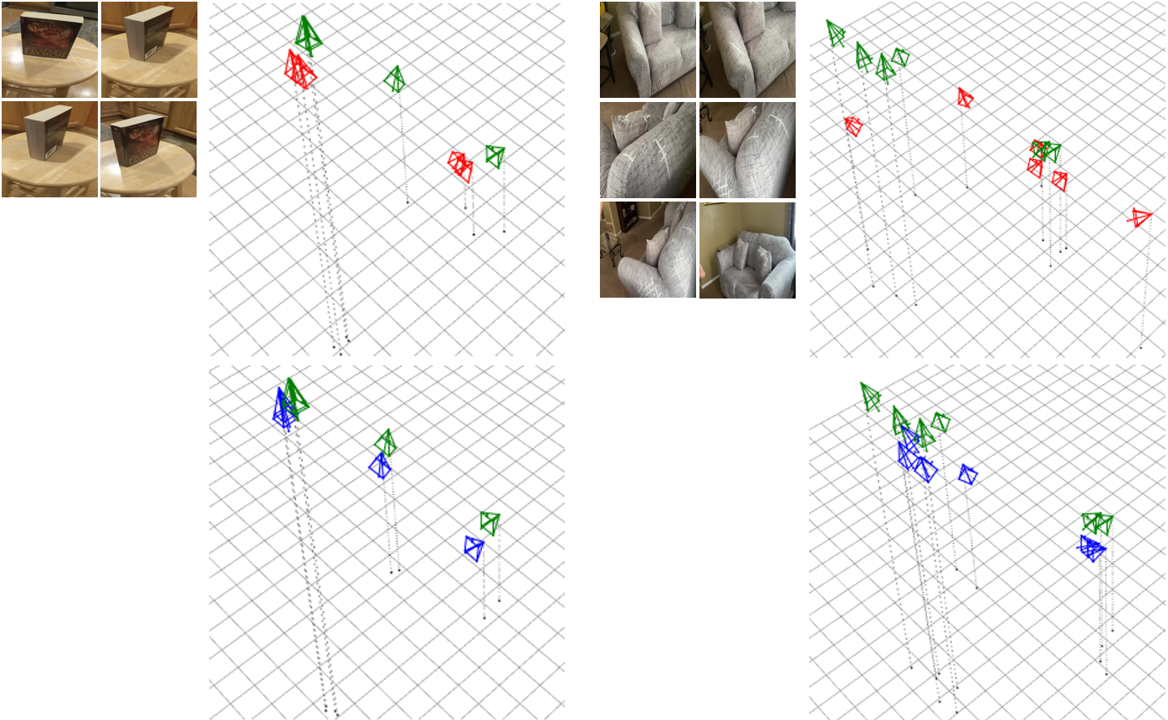}
  \includegraphics[width=1\linewidth]{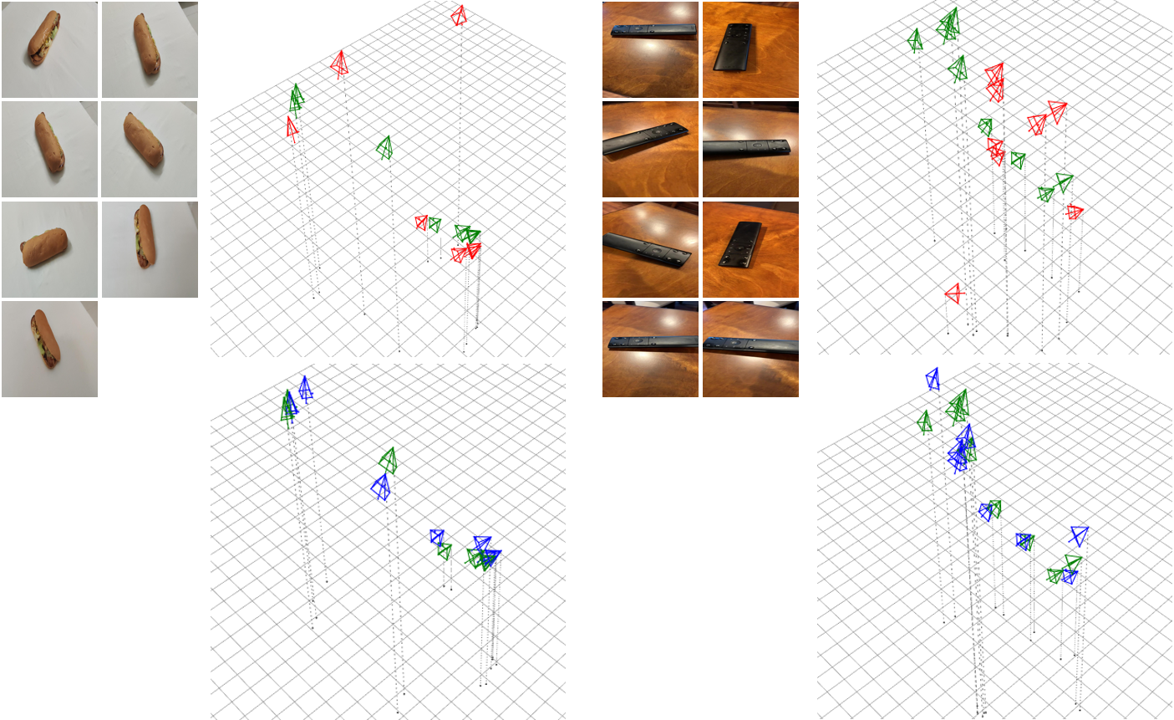}
  \caption{Visualization of recovered camera poses on CO3D. Ground truth, original predictions by Forge-2D, and refined poses with T-Graph (\textit{pair-t}) are shown in green, red, and blue, respectively.}
  \label{fig:co3d visual}
\end{figure}

To provide a more intuitive demonstration of the performance gains brought by introducing T-Graph to the baseline models, we visualized the recovered camera poses on several examples from both datasets. 
Specifically, Fig.~\ref{fig:co3d visual} compares the results of Forge-2D and the combination of Forge-2D with T-Graph(\textit{pair-t}) on the CO3D dataset. The visualization shows that the camera poses refined by T-Graph (blue cameras) are consistently closer to the ground truth (green cameras) than the original predictions (red cameras), in terms of both camera center positions and orientations.

\begin{figure}[H]
  \centering
  \includegraphics[width=1\linewidth]{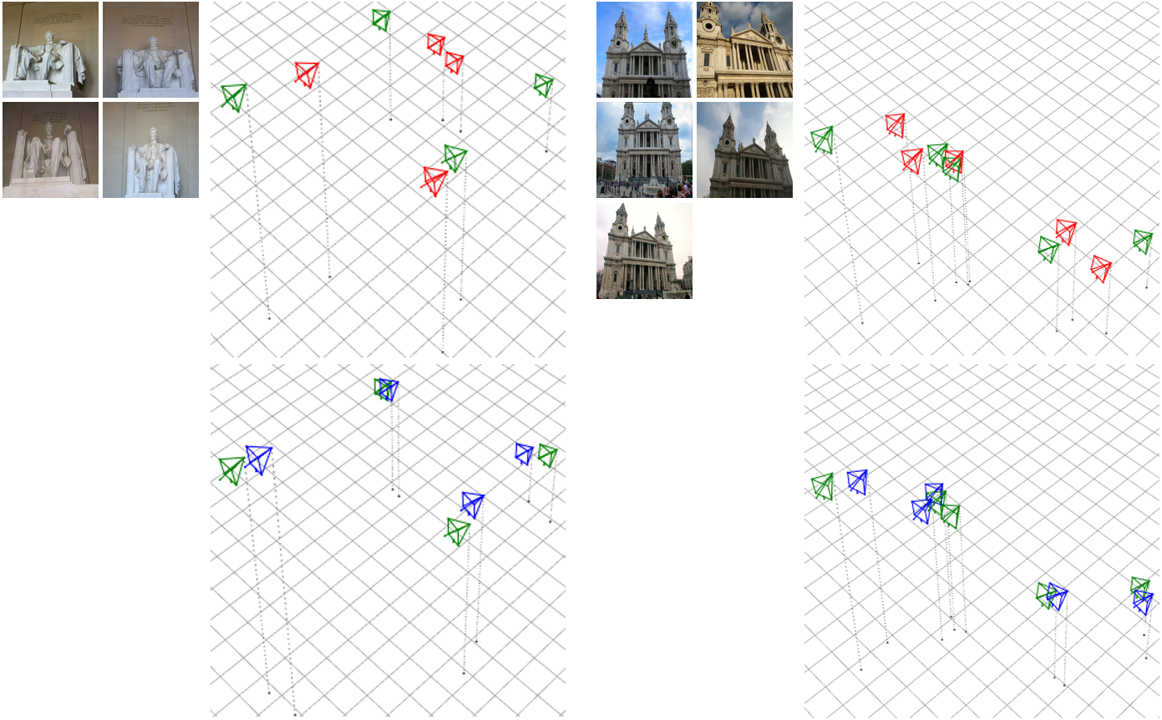}
  \includegraphics[width=1\linewidth]{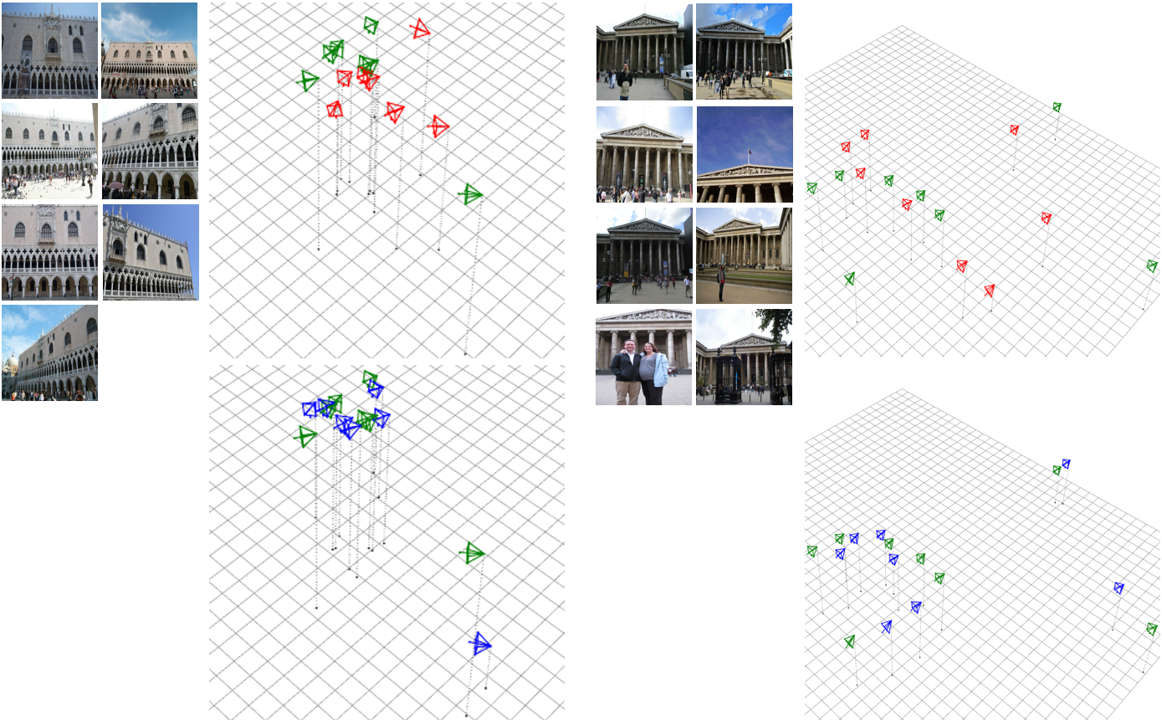}
  \caption{Visualization of recovered camera poses on IMC PhotoTourism. Ground truth, original predictions by RelPose++, and refined poses with T-Graph (\textit{relative-t}) are shown in green, red, and blue, respectively.}
  \label{fig:imc visual}
\end{figure}

Similarly, Fig.~\ref{fig:imc visual} presents the comparison between RelPose++ and RelPose++ with T-Graph(\textit{relative-t}) on the IMC PhotoTourism dataset. The scenes in this dataset are notably more challenging than those in CO3D, with significant variations in lighting conditions, occlusions from certain viewpoints, and substantial changes in camera-to-object distance. 
Despite these challenges, the visualizations clearly demonstrate that T-Graph improves the alignment of predicted camera poses with the ground truth, further confirming its effectiveness in enhancing prediction accuracy.

\section{Discussion}

Experimental results across four comparative settings consistently demonstrate that, regardless of the choice of pairwise translation representation, T-Graph effectively enhances the performance of the baseline model. Specifically, \textit{pair-t} proves to be better suited for the CO3D \cite{reizenstein2021common}, while \textit{relative-t} achieves superior results on the IMC Phototourism \cite{jin2021image}. To gain a clearer understanding of the differences between these two representations and their respective application scenarios, we categorize real-world camera configurations into three typical scenarios and analyze the characteristics of each, alongside the preferred choice of pairwise translation representations.

\begin{figure}[H]
  \centering
  \includegraphics[width=1\linewidth]{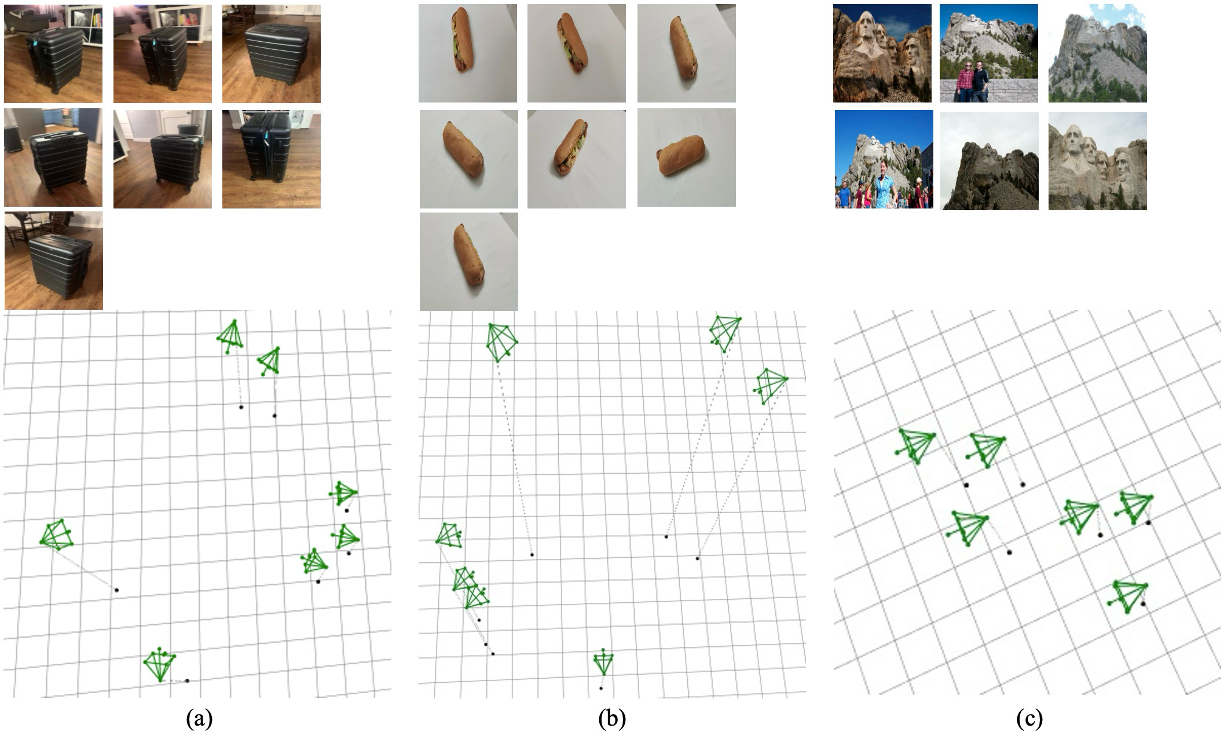}
  \caption{Camera pose visualization of different scenarios. (a) Center-facing cameras, (b) Mostly center-facing with a few parallel aligned cameras, (c) Mostly parallel aligned cameras.}
  \label{fig: case}
\end{figure}

In the first scenario, as illustrated in Fig.~\ref{fig: case}(a), cameras are arranged in a center-facing distribution around the target object, with their optical axes largely converging towards the same region. In this case, \textit{pair-t} provides an accurate description of the pairwise translation relationships, as the optical axes of each camera pair are nearly co-planar and exhibit clear convergence.

In the second scenario, as shown in Fig.~\ref{fig: case}(b), the majority of cameras still follow a center-facing distribution, but a minority of cameras are approximately parallel to each other (e.g., the two parallel cameras in the lower left corner). 
Since the center-facing configuration predominates and the influence of the parallel pairs is limited, \textit{pair-t} remains appropriate in this setting. 

In the third scenario, as depicted in Fig.~\ref{fig: case}(c), due to the increased distance between the cameras and the target object, most cameras tend to be approximately parallel to each other. In such cases, the estimation of intersection points between camera pairs becomes unstable, exhibiting substantial variance in their positions. Consequently, employing \textit{pair-t} to represent the pairwise translation relationships introduces learning difficulties for the model. Under these conditions, \textit{relative-t} is more effective.

In summary, the choice of the pairwise translation representation should be guided by the predominant characteristics observed across the dataset, particularly the spatial distribution of the cameras. 
For instance, \textit{pair-t} performs better on datasets with center-facing camera distributions (e.g., CO3D), while \textit{relative-t} is more effective for configurations with roughly parallel views, such as IMC Phototourism.

\section{Conclusion}
In this paper, we propose T-Graph, a lightweight and plug-and-play enhancement module to improve the performance of camera pose estimation models in sparse-view scenarios. 
T-Graph addresses a key limitation of existing end-to-end camera pose estimation methods, which often overlook pairwise translation information between viewpoints, hindering their performance. By explicitly modeling these pairwise relationships, T-Graph captures inter-camera correlations and enhances global structural awareness, resulting in notable performance improvements. Notably, T-Graph is only activated during training and does not change the inference process or efficiency of the baseline model at test phase. 
Furthermore, to accommodate various application scenarios, we introduce two distinct pairwise translation representations, \textit{relative-t} and \textit{pair-t}, supported by geometrical interpretation. 
Extensive comparative experiments demonstrate that the proposed T-Graph consistently benefits different baseline models and datasets, highlighting its effectiveness in improving camera pose estimation. 
Our findings also emphasize the importance of selecting an appropriate pairwise translation representation according to the characteristics of the camera-facing distribution in the dataset.
Moreover, we provide a novel perspective that may inspire future research: fully exploiting the information present in the ground truth, such as the translation relationships between pairwise viewpoints, offers a cost-effective approach to improve camera pose estimation performance.

\section*{Declaration of competing interest}
The authors declare that they have no known competing financial interests or personal relationships that could have appeared to influence the work reported in this paper.

\section*{Declaration of Generative AI and AI-assisted technologies in the writing process}
During the preparation of this work, the authors used ChatGPT (OpenAI) in order to improve the readability of the manuscript. After using this tool/service, the authors reviewed and edited the content as needed and take full responsibility for the content of the publication

\section*{Acknowledgements}
This work is financed by the Dutch Research Council NWO (www.nwo.nl) under the SUBLIME project (KICH1.ST01.20.008) of the NWO research programme KIC. It is also part of the Partnership Program of the Materials Innovation Institute M2i (www.m2i.nl) with project number N21007c. This work used the Dutch national e-infrastructure with the support of the SURF Cooperative using grant no. EINF-7472.



 \bibliographystyle{elsarticle-num} 





\end{document}